\documentclass[letterpaper, 10 pt, conference]{ieeeconf}  %

\IEEEoverridecommandlockouts                              %


\usepackage{booktabs}
\usepackage{makecell}
\usepackage{graphicx}
\usepackage{physics}
\usepackage{caption}
\usepackage{xspace}

\usepackage{url}\makeatletter\let\NAT@parse\undefined\makeatother\usepackage[numbers,sort&compress]{natbib}
\usepackage{array}
\usepackage{pifont}
\usepackage{xcolor}
\usepackage{cuted}   %
\usepackage{listings}
\usepackage{cleveref}
\newcolumntype{M}{>{\raggedright\arraybackslash}m{83pt}}    %
\newcolumntype{T}{>{\centering\arraybackslash}m{37.4pt}}    %
\newcolumntype{S}{>{\centering\arraybackslash}m{32pt}}      %
\newcommand{\hd}[1]{\footnotesize #1}                       %
\newcommand{\hdata}{\multicolumn{1}{@{}>{\centering\arraybackslash}m{68pt}|}{\hd{Data}}}
\newcommand{\cmark}{\ding{51}}
\newcommand{\xmark}{\ding{55}}

\newcommand{\algabbr}{DexAgent\xspace}
\newcommand{\up}{$\uparrow$}
\newcommand{\dn}{$\downarrow$}
\newcommand{\bsizeB}{\fontsize{7.6}{8.6}\selectfont}
\newcommand{\promptsec}[1]{\subsubsection{#1}\leavevmode\par\vspace{0.5ex}}

\title{\LARGE \bf
DexAgent: An Agentic Human2Sim2Robot Framework for Dexterous Manipulation with Self-Evolving Tool Library }

\author{Youhui Wang$^{1}$, Yunzhu Li$^{2}$, Li Fei-Fei$^{1}$, Jiajun Wu$^{1\dagger}$, Huang Huang$^{1\dagger}$\\[0.3em]
$^{1}$Stanford University \qquad $^{2}$Columbia University
}

\newcommand\blfootnote[1]{%
  \begingroup
  \renewcommand\thefootnote{}\footnote{#1}%
  \addtocounter{footnote}{-1}%
  \endgroup
}

\makeatletter\providecommand{\bstctlcite}[1]{\if@filesw\immediate\write\@auxout{\string\citation{#1}}\fi}\makeatother
\begin{document}\bstctlcite{IEEEauthorcontrol}

\thispagestyle{empty}
\pagestyle{empty}

\newcommand{\teaserCaption}{
\textbf{\algabbr.} An agentic Human2Sim2Robot framework that converts a single egocentric human video and a task prompt \textbf{(Left)} into robot training data. \algabbr uses property specific skills and verifiers from a self-evolving tool library \textbf{(Middle Bottom)} for each task. We evaluate \algabbr on 11 dexterous manipulation tasks spanning diverse object types and long-horizon interactions \textbf{(Right)}, achieving the highest success rates among the evaluated methods. 
}
\twocolumn[{
    \renewcommand\twocolumn[1][]{#1}
    \maketitle
    \centering
    \begin{minipage}{\linewidth}
        \centering
        \includegraphics[width=\linewidth]{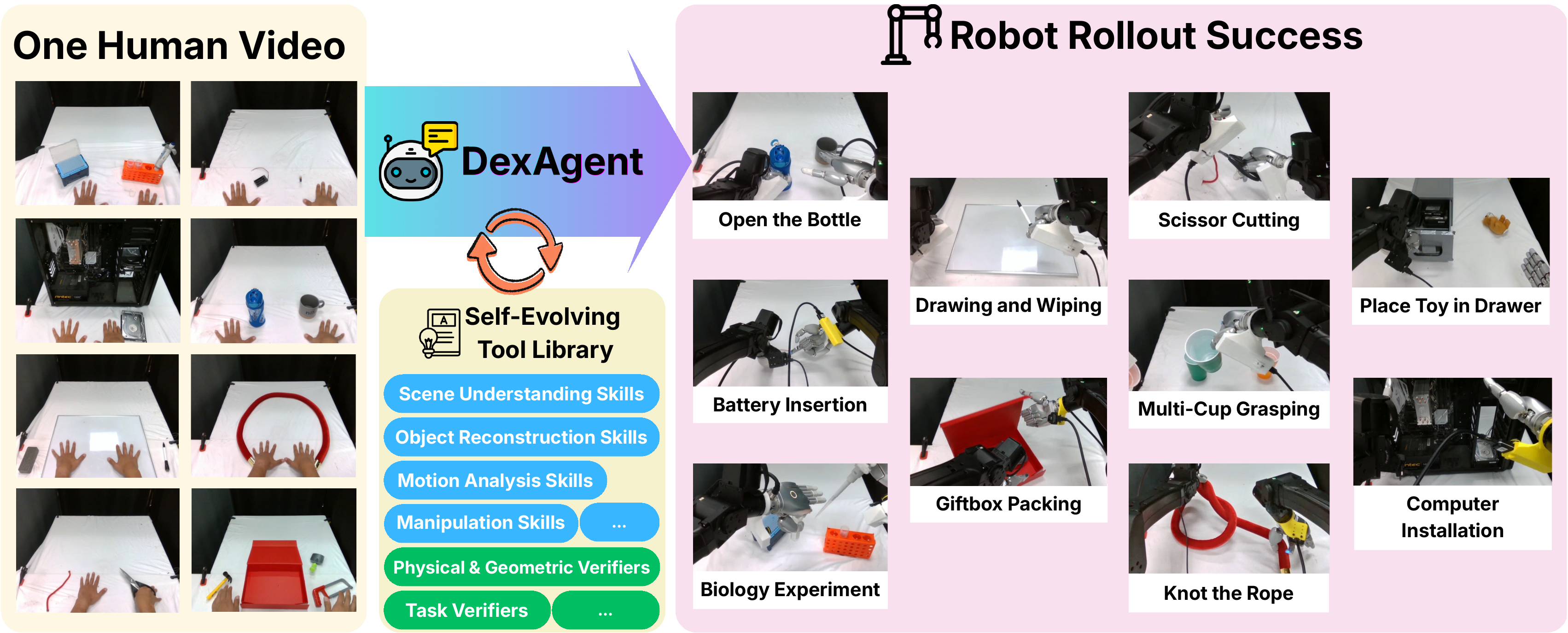}
    \end{minipage}
    {\captionsetup{hypcap=false}\captionof{figure}{\teaserCaption}\label{fig:teaser}}
    \vspace{0.5em}
}]
\blfootnote{$^{\dagger}$Equal advising.}
\blfootnote{\raggedright Corresponding author: Youhui Wang (\texttt{jeffreywang0303@cs.stanford.edu}).\par}

\begin{abstract}

Human videos offer a scalable source of demonstrations for dexterous robot manipulation. However, existing human-to-simulation-to-robot (Human2Sim2Robot) pipelines rely on predefined procedures that struggle to accommodate diverse object properties and interactions, particularly those involving articulated and deformable objects. We introduce \algabbr, an agentic Human2Sim2Robot framework that converts a single egocentric human video and a task prompt into physically grounded robot trajectories for policy training. It operates through four stages: semantic understanding of human videos, property-based simulation reconstruction, robot trajectory optimization, and robot data generation. At each stage, \algabbr adapts its approach to the task and object properties by selecting suitable skills from its tool library or developing new ones when needed. Property-specific verifiers assess stage outcomes for physical validity and task-specific requirements and provide feedback for refinement, preventing error propagation through the workflow. This adaptive, verification-guided process allows \algabbr to process diverse objects and long-horizon tasks. In the final stage, \algabbr varies object and robot states in simulation to generate diverse robot trajectories from a single human video, then retextures the rendered observations to facilitate sim-to-real transfer. Newly developed skills and verifiers are retained in its tool library, making it self-evolving to accumulate reusable capabilities. This reduces processing time as \algabbr encounters more human videos. Across eleven real-world tasks, policies trained with \algabbr-generated data achieve a 3.5$\times$ higher success rate than competing baselines. Project website: https://dexagent1.github.io/.

\end{abstract}

\section{INTRODUCTION}

Learning dexterous manipulation policies requires large amounts of robot data, yet collecting such data in the real world is expensive and difficult to scale, especially for multifingered robot hands. Human videos provide an abundant source of demonstrations for diverse object interactions, but converting them into physically valid robot trajectories remains challenging. Human-to-simulation-to-robot (Human2Sim2Robot) pipelines reconstruct human--object interactions in simulation to generate physically grounded robot trajectories. However, predefined reconstruction and motion-generation procedures struggle to accommodate diverse object properties and interactions. Articulated and deformable objects require different representations and constraints, while long-horizon tasks involve subgoals that may require different motion-generation strategies.

We introduce \algabbr, an agentic Human2Sim2Robot framework that converts a single egocentric human video and a task prompt into diverse, physically grounded robot trajectories for policy training. \algabbr adapts its four-stage workflow to task and object properties by selecting existing skills or developing new ones. First, \textbf{semantic understanding} identifies relevant object properties and decomposes the demonstration into subgoals. \textbf{Property-based simulation reconstruction} then builds a scene suited to the objects and interactions in the video. For \textbf{robot trajectory optimization}, the framework generates motion for each subgoal using human-motion-guided optimization or task-specific code. This flexibility supports long-horizon tasks involving rigid, articulated, and deformable objects. Finally, \textbf{robot data generation} augments verified trajectories by varying object and robot states in simulation and retextures the rendered observations for sim-to-real transfer. At each stage, property-specific verifiers assess physical validity and provide feedback for refinement. The agent iterates within the stage until all verifiers pass before advancing to the next stage, limiting error propagation from reconstruction to robot data generation. New skills and verifiers are retained in its tool library. As the framework processes more videos, it accumulates reusable capabilities, making it \textbf{self-evolving} and reducing the time to process subsequent demonstrations.

We evaluate \algabbr across eleven real-world tasks involving rigid, articulated, and deformable objects. Policies trained with \algabbr-generated data achieve a 3.5$\times$ higher success rate than competing baselines. We contribute:

1. \algabbr: an agentic Human2Sim2Robot framework that selects or develops property-specific skills to generate physically grounded robot trajectories from a single human video across diverse dexterous tasks.

2. A verification-guided generation process that iteratively refines each stage using simulation feedback verifiers, limiting error propagation throughout the pipeline.

3. A self-evolving library that retains newly developed skills and verifiers for reuse across videos, reducing processing time and supporting scalable data generation.

4. Physical experiments across eleven dexterous manipulation tasks with multifingered robot hands, demonstrating improved policy success over the evaluated baselines.

\section{RELATED WORK}

Dexterous manipulation requires coordinating many degrees of freedom through complex hand--object contacts. Reinforcement learning (RL) has enabled policies trained in simulation to transfer to physical robot hands, but exploration and task-specific reward design remain challenging~\cite{Lum2024DextrAH, Singh2024DextrAHRGB, openai2019learningdexterousinhandmanipulation}. Human demonstrations help address these challenges by providing motion priors, object trajectories, and contact information to guide policy learning~\cite{qin2021dexmv, mandikal2022dexviplearningdexterousgrasping, sharma2026demonstrationobjectsgeneralizingmanipulation, Mandi2025DexMachina, Li2025ManipTrans}. These methods often require high quality human priors to reduce exploration difficulty, and effective learning still depends on suitable reward formulations and training procedures. In contrast, \algabbr extracts human motion priors from video and combines task decomposition, trajectory optimization, and code-generated motion to produce robot trajectories without requiring task-specific RL reward design.

A complementary line of work combines human-data pretraining or co-training with robot demonstrations to learn manipulation representations and policies. R3M~\cite{nair2022r3m} learns visual representations from egocentric human videos for downstream robot imitation, while EgoVLA~\cite{yang2025egovlalearningvisionlanguageactionmodels} uses human-video pretraining to improve robot policy learning. EgoMimic~\cite{kareer2024egomimicscalingimitationlearning} co-trains policies on aligned human and robot data, and EgoScale~\cite{zheng2026egoscalescalingdexterousmanipulation} extends large-scale human-action pretraining to dexterous hands with limited robot demonstrations for adaptation. These approaches reduce robot-data requirements but still rely on robot demonstrations for training or adaptation. Collecting such demonstrations remains difficult to scale for multifingered hands.

To reduce reliance on robot teleoperation data, Human2Sim2Robot methods reconstruct human demonstrations in simulation and use physical interaction to bridge the embodiment gap. \citet{lum2025crossinghumanrobotembodimentgap} derive object-centric rewards and exploration guidance from a single RGB-D demonstration to learn transferable policies. Do as I Do~\cite{doasido} reconstructs hand--object interactions from monocular videos and generates robot trajectories through physics-aware optimization. SPIDER~\cite{spider} uses physics-based sampling to retarget and augment human motion, while TopoRetarget~\cite{toporetarget} preserves hand--object interactions to generate references for policy learning. EgoInfinity~\cite{egoinfinity} integrates reconstruction and retargeting for scalable video-to-action conversion, while V2D~\cite{v2d} combines agentic video ingestion with reconstruction and RL-based robotic grounding. However, these pipelines still struggle to extend to diverse object types and long-horizon tasks due to its fixed pipeline. \algabbr addresses this challenge using an agentic framework through property-specific skill selection and verification, choosing between human-motion-guided optimization and code-generated motion for each subgoal. Newly developed skills and verifiers are retained in a self-evolving library, allowing the framework to accumulate reusable capabilities across tasks.\looseness=-1

Agentic frameworks solve robotic tasks by generating and refining executable policies. CaP-X~\cite{fu2026capxframeworkbenchmarkingimproving} composes perception and control primitives into programs, improving them through execution feedback and automatic skill synthesis. GaP~\cite{chen2026gapgraphaspolicymultiagentselflearning} constructs graphs of modular robot skills and refines their structure and parameters through simulation. These approaches depend on reliable underlying skills, which are challenging to develop for multi-fingered robot hands. Instead, \algabbr applies an agentic framework to human-to-simulation transfer, deriving robot hand motions from human videos and refining them in simulation.\looseness=-1

\section{METHODOLOGY}

\begin{figure*}[t]
    \centering
    \includegraphics[width=1\linewidth]{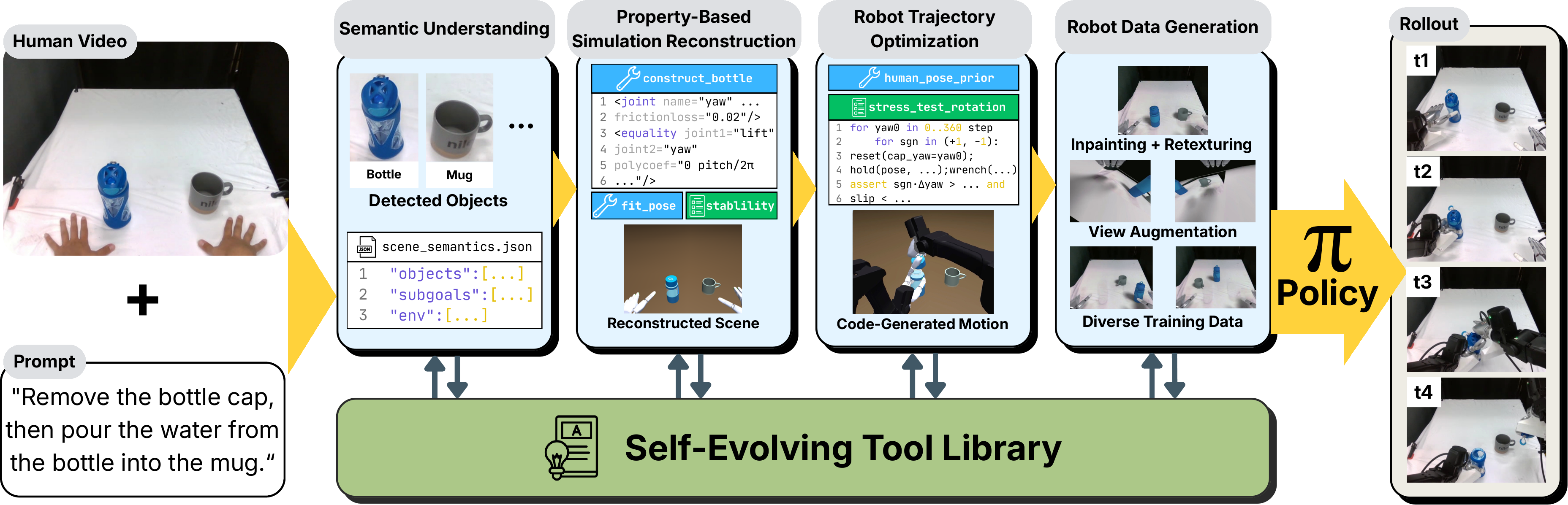}
    \caption{\textbf{DexAgent} converts a single egocentric human video and a task prompt into robot training data through four stages. \textbf{Stage 1} decomposes the task into subgoals and identifies the manipulated objects and their key properties. \textbf{Stage 2} reconstructs the objects in simulation using property-specific skills (blue) from the tool library, creating new skills when needed. \textbf{Stage 3} generates robot-hand trajectories by combining optimization guided by human motion priors with code-generated motion. \textbf{Stage 4} varies viewpoints and object states to produce diverse simulation data, inpaints the original human video, and retextures the augmented simulation videos for policy training. }
    \label{fig:method}
    \vspace{-0.5cm}
\end{figure*}

We propose \algabbr, an agentic Human2Sim2Robot framework for generating diverse physically valid robot trajectories from one human video. \algabbr processes each human video through 4 stages, shown in Fig.~\ref{fig:method}. Each stage takes the output of the previous stage and selects existing skills or develops new ones based on task properties. The agent evaluates the generated results using the stage's verifiers and uses any failure feedback to refine the results or revise its choice of skills. The generation--verification loop continues until all required verifiers pass or the execution budget is exhausted. Only passing outputs advance to the next stage. In stage 1, \algabbr perform semantic understanding to get the object properties involved in the task performed in the human video, and decompose the human motion into multiple subgoals. These properties are used to select skills and verifiers in later stages. In stage 2, \algabbr reconstruct the task in simulation by creating object asset. Then in stage 3, \algabbr generates robot trajectories for each decomposed subgoal through optimization based on extracted human motions or code generated trajectories. After generating one valid robot trajectory, \algabbr augments and retextures it to generate diverse robot trajectories for policy training for physical experiments in the final stage.

 Both skills and verifiers live in its tool library. New skills and verifies are saved into the tool library, making it self-evolving as it processes more human videos. Each stage invocation is either capped by a given budget or 100 agent turns, and a stage that runs out of its budget without passing the verifiers is recorded as a failure rather than being allowed to deliver with an incomplete result. The agent backbone can be any VLM swicthed between API.

\subsection{Skills and Verifiers}

The self-evolving tool library mainly stores two kinds of tools: \emph{Skill} and \emph{Verifier}. A \emph{skill} produces an asset, a pose or a motion. A \emph{verifier} accepts or rejects something a skill produced based on the ineraction inside simulation. Both skills and verifiers are ordinary Python functions with typed arguments. When improving a skill or a verifier, it is forked from the original function rather than overwriting on it. All the skills and verifiers are saved in the library after being created, and saving objects' assets can save execution time but is optional due to its size variance. Both skills and verifiers are designed to have high replicability and usability across samples, so the design of any tool does not overfit to any sample specifically. Skills and verifiers are mainly used for the following reconstruction stage and trajectory optimization stage, but the they are accessible throughout the process. The following sections provide more detailed explanation on the functions of specific skills and verifiers in each stage.\looseness=-1

\subsection{Semantic Understanding}

The goal of the first stage is to identify key objects in the task, the objects types and their key properties, and breakdown the entire task into multiple pairs of subgoals and the hands needed to execute the subtask. This breakdown sets clear goals for stage 2 reconstructon stage on what properties each object must have and for stage 3 trajectory optimization stage on what's the success definition of each subtask that the trajectory should optimize toward to.

Starting from the beginning, the input of \algabbr is one egocentric RGB human video performing the task and one natural-language task prompt. The VLM begins its inference with the language prompt, breaking down the prompt into possible objects in this entire task. Next, the agent infers the properties each object should potentially have, and bring such assumptions to the RGB human video. In the video, agent extracts frames and infer the objects' movements to confirm its property assumptions. For the properties that are confirmed, the agent records the key properties into a JSON file for later use. In addition to directly read from the figure, the agent has access to apply skills that can provide depth estimation, gravity directions, or human poses to further help it confirming the property extractions. In addition, the agent is required to decompose a task into each subtasks based on the demonstration, such that a subtask only includes one motion or as few as possible. For example, a pick-and-place task would be decomposed into 3 subtasks, which are 1) pick up the object, 2) move the object, and 3) place the object.

The output from this stage is a JSON file that incorporates 1) all the objects and their types, such as rigid, articulated, deformable, or others. 2) The property of each object that must be shown during reconstruction stage. 3) The subtasks decomposed from the entire task that includes which hand (or both) is the acting hand in the task and which object, and the subtask's goal. So far all the information are described in natural language but not parameters or values, and these are done in the following stages. The resulting JSON file is the asset sharable throughout the process.

\subsection{Property-Based Simulation Reconstruction}

The goal of Stage 2 is to reconstruct the objects that have the faithful properties and place in the simulation scene ready to be manipulated. With the object and task information extracted from the previous stage, \algabbr reconstructs the task in the simulation by creating and placing the object assets. 
Specifically, it first selects or writes new verifiers to each reconstructed object asset based on the object properties received from Stage 1. For example, for rigid objects, it checks the silhouette, scale, mass and resting-stability, for articulated object, it checks on joint-existence, axis, travel-range and self-locking, and for deformable object, it checks on topology. This ensures verifiers are based on what properties the object actually requires.

\algabbr then selects skills for reconstruction. If a skill for reconstructing objects from the category exists in the tool library, \algabbr calls this skill with specific arguments. Some arguments examples include the appearance (size, color, texture) and internal properties (number of joints, degree a joint can move). Otherwise, it writes one. A reconstruction skill is required to be written as a parameterized function for a group of objects with similar properties rather than a specific object. In other words, it must be reusable or replicable. When a reconstruction skill is first built, the agent begins with the editable URDF file of the object with no joints. Then, for each property retrieved from stage 1, the agent first defines the verifiers toward such property goal, then it starts editing inside URDF toward such property. Once the verifier is solved, DexAgent moves on to the next property. After all the properties with verifiers are solved, the agent will run all the verifiers jointly to ensure properties do not conflict. Once passes, the agent is required to conduct a second round check, in which it will start adding more new verifiers to test the reconstructed asset, and this round includes a test on generalizability that the paramalized function will reproduce 3 more variants and ensure those 3 variants also pass the verifiers. In that way, the reconstruction skill is considered usable and will be stored in the library. 

The verified object assets are then placed into the simulation environment with the robot setup. \algabbr finds the object placement by optimizing the IoU score between the object mask rendered in simulation and in human video from the same camera view. The robot setup and initial pose are given and fixed across samples, unless manually set to be different. Camera angle, extrinsics, and intrinsics are estimated mainly by depth for the distance and the overlay of sharpa hand and human hand for the camera angle, but it can also be manually set if such parameters are available. Stage 2 finally passes a scene file including the robot setup and the objects, where the placement and camera angle follows exactly the original video, which are ready to be manipulated. 

\begin{table}[t]
\centering
\small
\setlength{\tabcolsep}{2.8pt}
\caption{Reconstruction accuracy on HOI4D, scored separately for rigid and
articulated objects. Evaluation metrics are F-5 and F-10, which are F-scores at
two distance thresholds, and CD, which is Chamfer distance. Best entry in each
column in bold.}
\label{tab:hoi4d_recon}
\begin{tabular}{@{}l rrr rrr@{}}
\toprule
 & \multicolumn{3}{c}{Rigid} & \multicolumn{3}{c}{Articulated} \\
\cmidrule(lr){2-4} \cmidrule(lr){5-7}
Method & F-5\up & F-10\up & CD\dn & F-5\up & F-10\up & CD\dn \\
\midrule
HO~\cite{ho} & 0.28 & 0.51 & 3.86 & 0.29 & 0.47 & 1.30 \\
IHOI~\cite{ihoi} & 0.42 & 0.70 & 2.70 & 0.32 & 0.47 & 1.47 \\
HORSE~\cite{horse} & 0.26 & 0.45 & 6.69 & 0.19 & 0.34 & 1.91 \\
MCC-HO~\cite{mccho} & 0.52 & 0.78 & 1.36 & 0.35 & 0.55 & 1.21 \\
G-HOP~\cite{ghop} & 0.69 & 0.91 & 0.63 & 0.07 & 0.09 & 1.23 \\
FoundationPose~\cite{foundationpose} & 0.71 & 0.91 & 0.49 & 0.40 & 0.60 & 1.26 \\
Any6D~\cite{any6d} & 0.71 & 0.91 & 0.50 & 0.38 & 0.60 & 1.24 \\
Do as I Do~\cite{doasido} & 0.72 & 0.91 & 0.49 & 0.40 & 0.61 & 1.25 \\
\midrule
\textbf{\algabbr (ours)} & \textbf{0.83} & \textbf{0.96} & \textbf{0.29} & \textbf{0.47} & \textbf{0.68} & \textbf{1.08} \\
\bottomrule
\end{tabular}
\vspace{-2em}
\end{table}

\subsection{Trajectory Optimization}
In this stage, \algabbr generates robot trajectories that complete the task in the reconstructed simulation. For each subtask identified during semantic understanding, the agent first defines a task-specific success condition. For example, transporting an object requires reaching a target position and manipulating an articulated object requires reaching a target object joint state.%
These conditions allow the agent to evaluate whether the generated motion accomplishes the subtask.

\algabbr then selects a suitable skill from its library to determine the target robot hand pose for each subtask. It can directly retarget the demonstrated human pose, optimize the robot's finger joints to reproduce demonstrated hand--object contacts, or generate a grasp based on task requirements such as contact, force closure, and clearance. This flexibility enables the agent to use the human motion as a guide and adapt the grasp when the robot cannot reproduce it reliably.

Each target pose is evaluated using task-specific verifiers that evaluate whether it supports the required interaction. For example, a pose for transporting an object is tested through lifting and shaking and a pose for manipulating an articulated object is tested through sliding and spinning to ensure all the degrees a joint can reach are possible under the manipulation of such pose. If a pose fails verification, the agent uses that feedback to refine it or select another tool. Only poses that pass these checks are used for trajectory generation.

Given the verified target robot hand poses, inverse kinematics determines the corresponding arm configurations. The agent then generates code to connect these key poses, optionally using the demonstrated trajectory as a motion prior. Subtask trajectories are combined into a complete episode, with earlier motions revised when their resulting states prevent later subtasks from succeeding. Finally, whole-trajectory verifiers check motion limits and physical consistency before the trajectory is used for robot data generation.

\subsection{Data Generation and Policy Training}

In the final stage, \algabbr augments the verified robot trajectory in simulation to generate diverse policy-training data from a single human video. The agent varies the initial object and robot states within the reachable workspace and adapts the trajectory to each configuration. Each augmented episode passes the same task-specific pose and whole-trajectory checks as the original episode. Reusing the reconstructed assets and previously generated solution reduces the cost of producing additional demonstrations.

To reduce the visual gap between simulation and the real world, \algabbr removes the demonstrator's hands and manipulated objects from the source video using VOID~\cite{void}. The simulated robot and objects are rendered from the calibrated camera and composited into the cleared regions. Their appearance is further refined through retexturing in Blender~\cite{blender}. Additional camera views, including wrist views, can also be rendered in simulation to provide the observations required by the robot policy.
Together, trajectory augmentation and visual processing convert a single human demonstration into diverse, verified robot episodes for policy training.\looseness=-1

\begin{table}[t]
\centering
\small
\setlength{\tabcolsep}{3pt}
\caption{Human-to-robot retargeting quality on OakInk. Success is the share of
clips whose converted trajectory passes the physics check. $E_{pos}$ and
$E_{rot}$ are the residual position and rotation gap from the demonstration,
lower is better. The two indented rows ablate the strongest baseline.}
\label{tab:oakink_retarget}
\resizebox{\linewidth}{!}{%
\begin{tabular}{@{}l rrr@{}}
\toprule
Method & Success \%\up & $E_{pos}$ (m)\dn & $E_{rot}$ (rad)\dn \\
\midrule
Dex-retargeting~\cite{dexretargeting} & 28.6 &  0.08 & 0.62 \\
SPIDER~\cite{spider} (mjwp) & 71.4 & 0.04 & 0.57 \\
SPIDER~\cite{spider} (mjwp\_act) & 77.1 & 0.04 & 0.42 \\
Do-as-I-Do~\cite{doasido} (Sharpa hand) & 81.0 & \textbf{0.03} & 0.15 \\
\quad without transition reward & 79.0 & \textbf{0.03} & 0.14 \\
\quad annealed sampling only & 72.0 & 0.08 & 0.32 \\
\midrule
\textbf{\algabbr (ours)} & \textbf{85.7} & \textbf{0.03} & \textbf{0.12} \\
\bottomrule
\end{tabular}
}
\vspace{-0.5em}
\end{table}

\section{Data Quality Experiments}\label{sec:data_quality}
We evaluate the data quality generated by \algabbr. MuJoCo~\cite{mujoco} is used as the simulator for this set of experiments. Specifically we want to answer the following questions: \textit{Does an agentic framework achieve better reconstruction compared to the baselines?} and \textit{Does \algabbr generate higher-quality robot trajectories from human motion?}

\subsection{Human--Object Interaction Reconstruction}
We evaluate human--object interaction reconstruction from video on HOI4D~\cite{hoi4d}, reporting F-scores at two distance thresholds and Chamfer distance separately for rigid and articulated objects (\Cref{tab:hoi4d_recon}).
\algabbr achieves the best results across both object categories. Relative to the strongest baseline for each metric in \Cref{tab:hoi4d_recon}, \algabbr improves F-5 and F-10 by 15.3\% and 5.5\%, respectively, and reduces Chamfer distance by 40.8\% on rigid objects. Articulated objects remain more challenging for the baselines, and \algabbr achieves larger relative gains in F-5 and F-10 of 17.5\% and 11.5\%, respectively, alongside a 10.7\% reduction in Chamfer distance.
These results suggest that reconstruction benefits from iterative verification of the generated scene and adapting to object properties. \algabbr checks geometric and physical consistency and uses failure feedback to refine poses, revise joint models, or develop alternative reconstruction skills. This process continues until the verifiers pass.

\subsection{Robot Trajectory Generation Quality}
We evaluate the quality of robot trajectories generated from human hand motion on OakInk~\cite{oakink} (\Cref{tab:oakink_retarget}). We evaluate the generated trajectories through physics-based
simulation rollouts. Following Do-as-I-Do [14], a clip is counted
as successful if its mean object position error is below
$0.1\,\mathrm{m}$ and its mean object rotation error is below
$0.5\,\mathrm{rad}$. Success reports the fraction of clips
satisfying both criteria. The positional and rotational errors
measure the time-averaged deviation of the simulated object
pose from the demonstration.\looseness=-1

For SPIDER, MJWP denotes the MuJoCo Warp implementation without
contact guidance, while MJWP-act denotes the variant with
actuator-based contact guidance. Sharpa Wave denotes the robot
hand used by Do-as-I-Do, which was a built-in asset option.
\algabbr achieves the highest success rate, with a 5.8\% relative improvement over the strongest baseline. It also reduces rotational error by 20.0\% while matching its positional error. These results indicate \algabbr improves physical validity without sacrificing motion fidelity.
These gains are attributed to the flexibility of subtask-level motion generation. For each subtask, \algabbr determines a target robot hand pose and refines it using task-specific verifier feedback until it passes the required checks. It then generates motion connecting verified poses, using human-motion guided optimization or task-specific code as appropriate. This enables it to adapt portions of the demonstration to task requirements and the robot's kinematics.

\begin{figure}[t]
    \centering
    \includegraphics[width=1\linewidth]{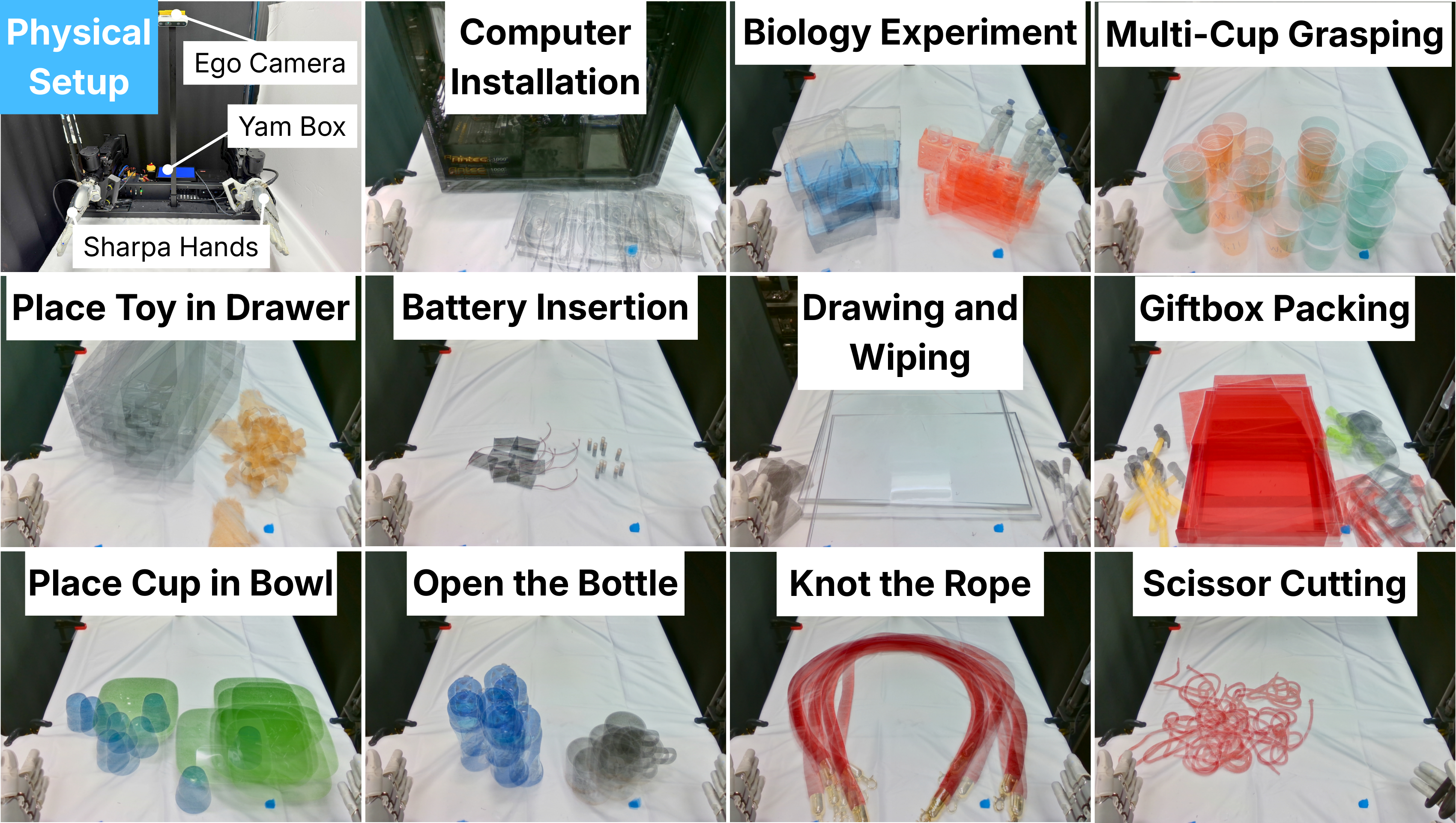}
    \caption{Initial object-state distributions for policy evaluation. Object placements are randomized across trials for each task.}
    \label{fig:evaldist}
    \vspace{-2em}
\end{figure}

\begin{figure}[th!]
    \centering
    \includegraphics[width=1\linewidth]{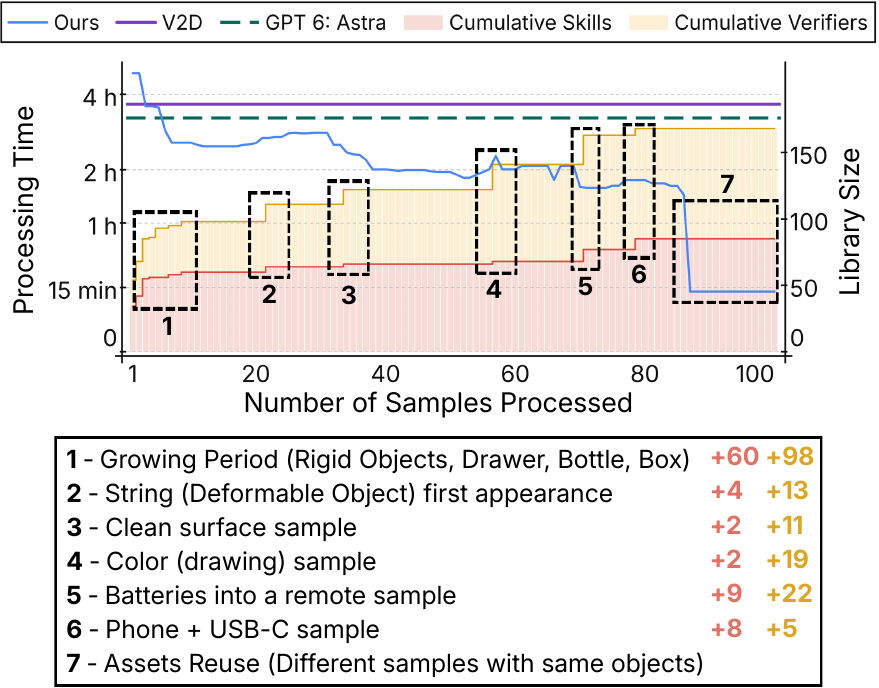}
    \caption{Processing cost falls as the library grows. Over 100 EgoDex
    samples the library reaches 85 skills and 168 verifiers. In terms of cost, V2D converting one sample takes \textbf{3.7 hours} on average and GPT 6: Astra takes \textbf{3.3 hours} on average at a flat rate, whereas \algabbr takes \textbf{2.1 hours} on average in a fresh run, a \textbf{36.4\%} reduction comparing to GPT 6: Astra and a \textbf{43.2\%} reduction comparing to V2D.
    }
    \label{fig:lifelong}
    \vspace{-2em}
\end{figure}

\begin{figure*}[th]
    \centering
    \includegraphics[width=1\linewidth]{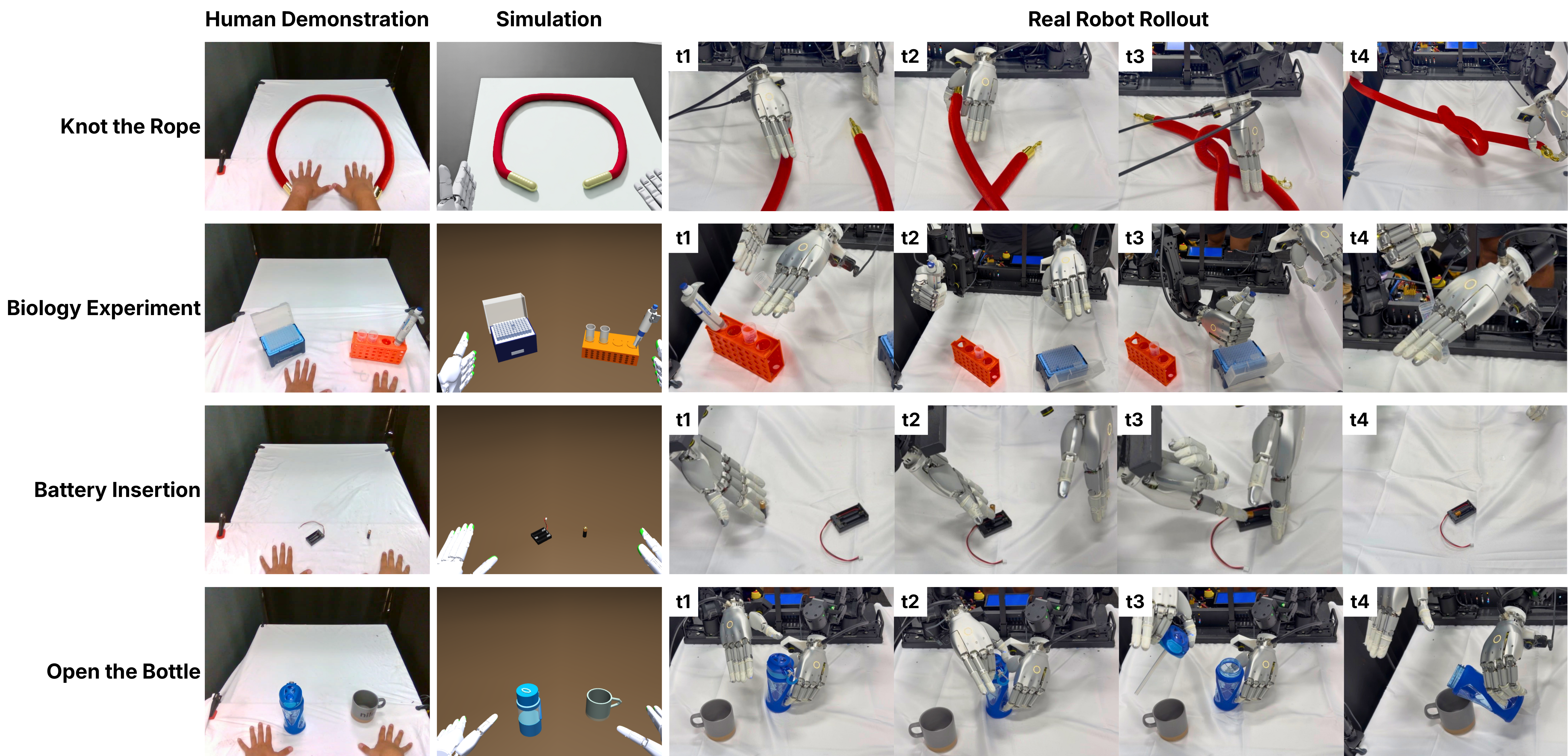}
    \caption{Real experiments results on four selected tasks. Each row shows the original human demonstration (\textbf{Column 1}), its reconstructed simulation scene(\textbf{Column 2}), and representative frames from the resulting real-robot rollout (\textbf{Columns 3--6}). The examples span diverse manipulation behaviors, including deformable-object manipulation, multi-object interaction, precise insertion, and articulated object manipulation.}
    \label{fig:rollout}
    \vspace{-2em}
\end{figure*}

\begin{table*}[t]
\centering
\caption{Real-world replay success across eleven tasks. Each converted
trajectory is executed open loop, which scores the conversion rather than a
learned policy. A task is marked successful if at least one of ten replay trials succeeds. GPT-6 Astra joined the benchmark after the first evaluation
round.}
\vspace{-0.1cm}
\label{tab:real_exp}
\small
\setlength{\tabcolsep}{0.5pt}
\resizebox{\linewidth}{!}{%
\begin{tabular}{@{}M | *{11}{T} | S@{}}
\toprule
\hdata & \hd{Cup} & \hd{Giftbox} & \hd{Drawer} & \hd{Rope Knot} & \hd{Scissors} & \hd{Drawing} & \hd{Computer} & \hd{Bottle} & \hd{Biology} & \hd{Battery} & \hd{Multi-object} & \hd{Total} \\
\midrule
Dex-retargeting~\cite{dexretargeting} & \cmark & \xmark & \xmark & \xmark & \xmark & \xmark & \xmark & \xmark & \xmark & \xmark & \xmark & 1/11 \\
Do as I Do~\cite{doasido} & \cmark & \xmark & \xmark & \xmark & \xmark & \xmark & \xmark & \xmark & \xmark & \xmark & \xmark & 1/11 \\
Spider~\cite{spider} & \cmark & \cmark & \cmark & \xmark & \cmark & \xmark & \cmark & \xmark & \xmark & \cmark & \xmark & 6/11 \\
TopoRetarget~\cite{toporetarget} & \cmark & \cmark & \cmark & \xmark & \xmark & \xmark & \xmark & \xmark & \xmark & \xmark & \xmark & 3/11 \\
Egoinfinity~\cite{egoinfinity} & \cmark & \xmark & \xmark & \xmark & \xmark & \xmark & \xmark & \xmark & \xmark & \xmark & \xmark & 1/11 \\
V2D~\cite{v2d} & \cmark & \cmark & \cmark & \xmark & \cmark & \xmark & \xmark & \xmark & \xmark & \xmark & \cmark & 5/11 \\
GPT-6: Astra~\cite{gpt6astra} & \cmark & \cmark & \cmark & \xmark & \xmark & \cmark & \cmark & \xmark & \xmark & \xmark & \xmark & 5/11 \\
\textbf{\algabbr (ours)}   & \textbf{\cmark} & \textbf{\cmark} & \textbf{\cmark} & \textbf{\cmark} & \textbf{\cmark} & \textbf{\cmark} & \textbf{\cmark} & \textbf{\cmark} & \textbf{\cmark} & \textbf{\cmark} & \textbf{\cmark} & \textbf{11/11} \\
\bottomrule
\end{tabular}
}

\vspace{0.5em}
\centering
\caption{Real-world success rate on the same task set, measured as the percentage of ten closed-loop trials per task. Each policy is trained on the data converted and augmented by the same protocol. The average is over all eleven tasks.}
\vspace{-0.1cm}
\label{tab:real_exp_policy}
\small
\setlength{\tabcolsep}{0.5pt}
\resizebox{\linewidth}{!}{%
\begin{tabular}{@{}M | *{11}{T} | S@{}}
\toprule
\hdata & \hd{Cup} & \hd{Giftbox} & \hd{Drawer} & \hd{Rope Knot} & \hd{Scissors} & \hd{Drawing} & \hd{Computer} & \hd{Bottle} & \hd{Biology} & \hd{Battery} & \hd{Multi-object} & \hd{Average} \\
\midrule
Dex-retargeting~\cite{dexretargeting} & 0\% & 0\% & 0\% & 0\% & 0\% & 0\% & 0\% & 0\% & 0\% & 0\% & 0\% & 0.0\% \\
Do as I Do~\cite{doasido} & 50\% & 50\% & 10\% & 0\% & 0\% & 0\% & 10\% & 0\% & 0\% & 0\% & 0\% & 10.9\% \\
Spider~\cite{spider} & 70\% & 40\% & 10\% & 0\% & 30\% & 20\% & 20\% & 0\% & 0\% & 10\% & 0\% & 18.2\% \\
TopoRetarget~\cite{toporetarget} & 60\% & 10\% & 0\% & 0\% & 10\% & 10\% & 0\% & 0\% & 0\% & 0\% & 0\% & 8.2\% \\
Egoinfinity~\cite{egoinfinity} & 60\% & 20\% & 0\% & 0\% & 0\% & 0\% & 0\% & 0\% & 0\% & 0\% & 0\% & 7.3\% \\
V2D~\cite{v2d} & 70\% & 40\% & 30\% & 0\% & 0\% & 10\% & 10\% & 0\% & 0\% & 0\% & 30\% & 17.3\% \\
GPT-6: Astra~\cite{gpt6astra} & 60\% & 40\% & 50\% & 0\% & 0\% & 30\% & 0\% & 0\% & 0\% & 0\% & 0\% & 16.4\% \\
\textbf{\algabbr (ours)}   & \textbf{90\%} & \textbf{70\%} & \textbf{80\%} & \textbf{50\%} & \textbf{80\%} & \textbf{40\%} & \textbf{50\%} & \textbf{70\%} & \textbf{40\%} & \textbf{60\%} & \textbf{70\%} & \textbf{63.6\%} \\
\bottomrule
\end{tabular}
}
\vspace{-0.3cm}
\end{table*}

\section{Physical Experiments}
We evaluate whether data generated by \algabbr support successful real-world dexterous manipulation. For each task, we record one complete human demonstration using RealSense D435, capturing an egocentric RGB view of the demonstrator's hands and manipulated objects. Videos are recorded at 640x480 and 30 frames per second. Each video captures a complete execution of the corresponding task and is paired with a natural-language prompt describing the task. For trajectory-replay evaluation, all methods receive the same demonstration video for each task. DexAgent uses the video and task prompt to reconstruct the scene in simulation and generate a verified robot trajectory. Additional robot demonstrations for policy training are produced through simulation-based augmentation of this trajectory. Depending on the task, the framework uses MuJoCo~\cite{mujoco} or Isaac Sim~\cite{isaacsim}. \looseness=-1

\subsection{Evaluation Setup}
All experiments are conducted on a bimanual YamBox station with two Sharpa hands, an ego-view RealSense D435 camera, and two wrist-mounted Zed Mini cameras.
We evaluate both trajectory replay and learned-policy performance over ten trials per task. For \textbf{replay evaluation}, each method receives the same human video and generates a trajectory that is executed ten times on the physical robot. A task counts as successful if at least one trial succeeds, measuring whether the generated robot trajectory is physically executable. For \textbf{policy evaluation}, we generate 500 robot episodes with one ego-view and two wrist-view observations, and use them to fine-tune $\pi_{0.5}$~\cite{pi05}. Each trained policy is tested over ten trials with randomized initial object states (\Cref{fig:evaldist}), and we report the fraction of successful trials.\looseness=-1

The baselines address different parts of the Human2Sim2Robot pipeline: some focus on reconstruction, others on motion retargeting, and some support both. To enable a fair end-to-end comparison, we retain each method's supported components and supply a common implementation for the missing stages. Specifically, Dex-retargeting~\cite{dexretargeting}, SPIDER~\cite{spider}, and TopoRetarget~\cite{toporetarget} retain their retargeting procedures and receive the scene assets and object poses they require. For Do as I Do~\cite{doasido}, EgoInfinity~\cite{egoinfinity}, and V2D~\cite{v2d}, we retain their reconstruction and trajectory-generation components where supported. We then apply the same data-augmentation and visual-processing protocol to all methods and use the resulting data to train the same policy model. This setup compares each method's contribution while controlling the remaining stages of the pipeline. GPT-6: Astra~\cite{gpt6astra} is prompted zero-shot to generate robot trajectories without our harness and evaluated under the same protocol.\looseness=-1

We evaluate on eleven dexterous tasks: cup placement, multi-cup grasping, and battery insertion test rigid-object manipulation; bottle opening, drawer placement, and scissor cutting involve articulated objects; and rope knotting tests deformable-object manipulation. Drawing and wiping and pipetting test tool use, while gift-box packing and computer installation require long-horizon coordination.

\subsection{Results Analysis}
 We aim to answer the following questions.

\textit{Does \algabbr generate more physically executable robot trajectories from human videos?}
\Cref{tab:real_exp} reports successful replay on all eleven tasks for \algabbr, compared with six for the strongest baseline, SPIDER, suggesting that the robot trajectories generated by \algabbr is more physically executable and \algabbr works on different tasks type, while other baselines struggle on tasks with deformable, articulated or long horizon tasks.

\textit{Do policies trained with \algabbr-converted data perform better in physical environments?}
As shown in~\Cref{sec:data_quality}, \algabbr generates higher quality robot trajectories. We study whether this helps robot policies. As shown in~\Cref{tab:real_exp_policy}, \algabbr achieves an average success rate of 63.6\%, approximately 3.5$\times$ that of SPIDER and 3.7$\times$ that of V2D. Gains extend to articulated and deformable objects: drawer success reaches 80\%, compared with the highest reported baseline result of 50\%, while rope-knot success reaches 50\% and baselines score zero. \Cref{fig:rollout} shows representative successful robot rollouts. This highlights that \algabbr is a general framework applicable to a diverse, dexterous tasks.\looseness=-1

\textit{How does the tool library evolve with more human videos?}
\label{sec:lifelong}
\Cref{fig:lifelong} tracks \algabbr as it processes 100 EgoDex~\cite{egodex} videos. We randomly order the videos, placing recurring-object cases last to evaluate asset reuse. Starting from a base skill set with no task-specific tools, \algabbr develops and retains new skills and verifiers as it processes each video. By sample 78, its self-evolving library contains 85 skills and 168 verifiers, which subsequent videos can reuse alongside novel assets when objects recur. 

Reuse reduces processing time from 4.3 h for the first video to 2.2 h with skill reuse and 13.1 min with asset reuse. The full run produces 100 verified episodes in 210~h. V2D and GPT-6 Astra do not retain a persistent tool library, so their processing costs remain approximately 3.7~h and 3.3~h per episode. With its tool library and asset reuse, \algabbr produces a verified episode approximately 17 times faster than V2D and 15.1 times faster than GPT-6 Astra.\par Together, these results show that \algabbr supports physically executable trajectories and effective policy learning across diverse dexterous tasks. Its self-evolving library accumulates reusable skills and verifiers, reducing processing time for subsequent videos, with further savings when object assets can also be reused.

\section{CONCLUSION}

We introduced \algabbr, an agentic Human2Sim2Robot framework that converts a single egocentric human video into robot trajectories and policy-training data. The agent selects or develops task-specific skills and iterates each stage until all its verifiers pass. Its self-evolving library retains skills and verifiers for reuse, enabling capabilities to accumulate across videos. Experiments demonstrate improved data quality and real-world policy performance across eleven dexterous tasks, with lower processing costs as the library grows.

Remaining limitations include semantic and geometric errors, incomplete verification, and simulation-to-reality gaps. Undetected errors propagate through the library, and force-sensitive interactions can fail despite passing simulation checks. Future work will strengthen verification and explore reinforcement learning as a fallback when existing skills cannot produce a valid solution.

\section*{ACKNOWLEDGMENT}

We thank Sharpa for equipment support. We also thank Peter Kulits, Žiga Kovačič, Ziyu Chen, Chongkai Gao, and Jeff Tan for their meaningful discussions and help, and the entire Stanford SVL community for their continuous support.

\bibliographystyle{IEEEtranN}
\bibliography{references}

@inproceedings{Lum2024DextrAH,
title = {{DextrAH-G}: Pixels-to-Action Dexterous Arm-Hand Grasping with Geometric Fabrics},
author = {Lum, Tyler Ga Wei and Matak, Martin and Makoviychuk, Viktor and Handa, Ankur and Allshire, Arthur and Hermans, Tucker and Ratliff, Nathan D. and Van Wyk, Karl},
booktitle = {Proc.\ Conference on Robot Learning (CoRL)},
year = {2024}
}

@inproceedings{Li2025ManipTrans,
      title={ManipTrans: Efficient Dexterous Bimanual Manipulation Transfer via Residual Learning}, 
      author={Kailin Li and Puhao Li and Tengyu Liu and Yuyang Li and Siyuan Huang},
      year={2025},
      eprint={2503.21860},
      archivePrefix={arXiv},
      primaryClass={cs.RO},
      url={https://arxiv.org/abs/2503.21860}, 
    }

@misc{Mandi2025DexMachina,
      title={DexMachina: Functional Retargeting for Bimanual Dexterous Manipulation}, 
      author={Zhao Mandi and Yifan Hou and Dieter Fox and Yashraj Narang and Ajay Mandlekar and Shuran Song},
      year={2025},
      eprint={2505.24853},
      archivePrefix={arXiv},
      primaryClass={cs.RO},
      url={https://arxiv.org/abs/2505.24853}
}

@misc{Singh2024DextrAHRGB,
      title={DextrAH-RGB: Visuomotor Policies to Grasp Anything with Dexterous Hands}, 
      author={Ritvik Singh and Arthur Allshire and Ankur Handa and Nathan Ratliff and Karl Van Wyk},
      year={2025},
      eprint={2412.01791},
      archivePrefix={arXiv},
      primaryClass={cs.RO},
      url={https://arxiv.org/abs/2412.01791}, 
}

@misc{lum2025crossinghumanrobotembodimentgap,
      title={Crossing the Human-Robot Embodiment Gap with Sim-to-Real RL using One Human Demonstration}, 
      author={Tyler Ga Wei Lum and Olivia Y. Lee and C. Karen Liu and Jeannette Bohg},
      year={2025},
      eprint={2504.12609},
      archivePrefix={arXiv},
      primaryClass={cs.RO},
      url={https://arxiv.org/abs/2504.12609}, 
}

@misc{openai2019learningdexterousinhandmanipulation,
      title={Learning Dexterous In-Hand Manipulation}, 
      author={OpenAI and Marcin Andrychowicz and Bowen Baker and Maciek Chociej and Rafal Jozefowicz and Bob McGrew and Jakub Pachocki and Arthur Petron and Matthias Plappert and Glenn Powell and Alex Ray and Jonas Schneider and Szymon Sidor and Josh Tobin and Peter Welinder and Lilian Weng and Wojciech Zaremba},
      year={2019},
      eprint={1808.00177},
      archivePrefix={arXiv},
      primaryClass={cs.LG},
      url={https://arxiv.org/abs/1808.00177}, 
}

@misc{qin2021dexmv,
      title={DexMV: Imitation Learning for Dexterous Manipulation from Human Videos}, 
      author={Yuzhe Qin and Yueh-Hua Wu and Shaowei Liu and Hanwen Jiang and Ruihan Yang and Yang Fu and Xiaolong Wang},
      year={2022},
      eprint={2108.05877},
      archivePrefix={arXiv},
      primaryClass={cs.LG},
      url={https://arxiv.org/abs/2108.05877}, 
}

@misc{mandikal2022dexviplearningdexterousgrasping,
      title={DexVIP: Learning Dexterous Grasping with Human Hand Pose Priors from Video}, 
      author={Priyanka Mandikal and Kristen Grauman},
      year={2022},
      eprint={2202.00164},
      archivePrefix={arXiv},
      primaryClass={cs.RO},
      url={https://arxiv.org/abs/2202.00164}, 
}

@IEEEtranBSTCTL{IEEEauthorcontrol,
  CTLuse_forced_etal = {yes},
  CTLmax_names_forced_etal = {6},
  CTLnames_show_etal = {1}
}

@misc{gpt6astra,
  author = {{OpenAI}},
  title = {{GPT-6 Astra}},
  year = {2026}
}

@inproceedings{mujoco,
  author = {Todorov, E. and Erez, T. and Tassa, Y.},
  title = {{MuJoCo}: A physics engine for model-based control},
  booktitle = {Proc. IEEE/RSJ Int. Conf. Intelligent Robots and Systems (IROS)},
  year = {2012}
}

@misc{isaacsim,
  author = {{NVIDIA}},
  title = {{Isaac Sim}: Robotics simulation and synthetic data generation},
  year = {2024}
}

@misc{blender,
  author = {{Blender Online Community}},
  title = {{Blender}: A {3D} modelling and rendering package},
  howpublished = {Blender Foundation}
}

@misc{void,
      title={VOID: Video Object and Interaction Deletion}, 
      author={Saman Motamed and William Harvey and Benjamin Klein and Luc Van Gool and Zhuoning Yuan and Ta-Ying Cheng},
      year={2026},
      eprint={2604.02296},
      archivePrefix={arXiv},
      primaryClass={cs.CV},
      url={https://arxiv.org/abs/2604.02296}, 
}

@article{pi05,
  author = {{Physical Intelligence}},
  title = {{$\pi_{0.5}$}: A vision-language-action model with open-world generalization},
  journal = {arXiv preprint arXiv:2504.16054},
  year = {2025}
}

@inproceedings{hoi4d,
  author = {Liu, Y. and others},
  title = {{HOI4D}: A {4D} egocentric dataset for category-level human-object interaction},
  booktitle = {Proc. IEEE/CVF Conf. Computer Vision and Pattern Recognition (CVPR)},
  year = {2022}
}

@inproceedings{oakink,
  author = {Yang, L. and others},
  title = {{OakInk}: A large-scale knowledge repository for understanding hand-object interaction},
  booktitle = {Proc. IEEE/CVF Conf. Computer Vision and Pattern Recognition (CVPR)},
  year = {2022}
}

@article{egodex,
  author = {Hoque, R. and others},
  title = {{EgoDex}: Learning dexterous manipulation from large-scale egocentric video},
  journal = {arXiv preprint arXiv:2505.11709},
  year = {2025}
}

@inproceedings{ho,
  author = {Hasson, Y. and others},
  title = {Learning joint reconstruction of hands and manipulated objects},
  booktitle = {Proc. IEEE/CVF Conf. Computer Vision and Pattern Recognition (CVPR)},
  year = {2019}
}

@inproceedings{ihoi,
  author = {Ye, Y. and Gupta, A. and Tulsiani, S.},
  title = {What's in your hands? {3D} reconstruction of generic objects in hands},
  booktitle = {Proc. IEEE/CVF Conf. Computer Vision and Pattern Recognition (CVPR)},
  year = {2022}
}

@inproceedings{horse,
  author = {Prakash, A. and others},
  title = {{3D} reconstruction of objects in hands without real world {3D} supervision},
  booktitle = {Proc. European Conf. Computer Vision (ECCV)},
  year = {2024}
}

@article{mccho,
  author = {Wu, J. and others},
  title = {Reconstructing hand-held objects in {3D}},
  journal = {arXiv preprint arXiv:2404.06507},
  year = {2024}
}

@inproceedings{ghop,
  author = {Ye, Y. and others},
  title = {{G-HOP}: Generative hand-object prior for interaction reconstruction and grasp synthesis},
  booktitle = {Proc. IEEE/CVF Conf. Computer Vision and Pattern Recognition (CVPR)},
  year = {2024}
}

@inproceedings{foundationpose,
  author = {Wen, B. and others},
  title = {{FoundationPose}: Unified {6D} pose estimation and tracking of novel objects},
  booktitle = {Proc. IEEE/CVF Conf. Computer Vision and Pattern Recognition (CVPR)},
  year = {2024}
}

@misc{any6d,
      title={Any6D: Model-free 6D Pose Estimation of Novel Objects}, 
      author={Taeyeop Lee and Bowen Wen and Minjun Kang and Gyuree Kang and In So Kweon and Kuk-Jin Yoon},
      year={2025},
      eprint={2503.18673},
      archivePrefix={arXiv},
      primaryClass={cs.CV},
}

@article{doasido,
  author = {Paliwal, B. and Etukuru, H. and Liang, W. and Abbeel, P. and Shafiullah, N. M. M. and Malik, J.},
  title = {{Do as I Do}: Dexterous manipulation data from everyday human videos},
  journal = {arXiv preprint arXiv:2606.19333},
  year = {2026}
}

@article{spider,
  author = {Pan, C. and Wang, C. and Qi, H. and Liu, Z. and Bharadhwaj, H. and Sharma, A. and Wu, T. and Shi, G. and Malik, J. and Hogan, F.},
  title = {{SPIDER}: Scalable physics-informed dexterous retargeting},
  journal = {arXiv preprint arXiv:2511.09484},
  year = {2025}
}

@article{toporetarget,
  author = {Wu, J. and Yao, S. and He, G. and Liu, X. and Zeng, Z. and Jiang, X. and Yang, H. and Zhang, W. and Zhao, H.},
  title = {{TopoRetarget}: Interaction-preserving retargeting for dexterous manipulation},
  journal = {arXiv preprint arXiv:2606.16272},
  year = {2026}
}

@article{egoinfinity,
  author = {Wang, G. and others},
  title = {{EgoInfinity}: A web-scale {4D} hand-object interaction data engine for any-view robot retargeting and video-to-action robot learning},
  journal = {arXiv preprint arXiv:2606.17385},
  year = {2026}
}

@misc{v2d,
  author = {{NVIDIA Isaac}},
  title = {{Video to Data}: A pipeline from human demonstration video to robot-ready training data},
  url = {https://github.com/nvidia-isaac/video_to_data},
  year = {2026}
}

@inproceedings{dexretargeting,
  author = {Qin, Y. and others},
  title = {{AnyTeleop}: A general vision-based dexterous robot arm-hand teleoperation system},
  booktitle = {Proc. Robotics: Science and Systems (RSS)},
  year = {2023}
}

@misc{yang2025egovlalearningvisionlanguageactionmodels,
  title = {{EgoVLA}: Learning Vision-Language-Action Models from Egocentric Human Videos},
  author = {Ruihan Yang and Qinxi Yu and Yecheng Wu and Rui Yan and Borui Li and An-Chieh Cheng and Xueyan Zou and Yunhao Fang and Xuxin Cheng and Ri-Zhao Qiu and Hongxu Yin and Sifei Liu and Song Han and Yao Lu and Xiaolong Wang},
  year = {2025},
  eprint = {2507.12440},
  archivePrefix = {arXiv},
  primaryClass = {cs.RO},
  url = {https://arxiv.org/abs/2507.12440}
}

@misc{kareer2024egomimicscalingimitationlearning,
  title = {{EgoMimic}: Scaling Imitation Learning via Egocentric Video},
  author = {Simar Kareer and Dhruv Patel and Ryan Punamiya and Pranay Mathur and Shuo Cheng and Chen Wang and Judy Hoffman and Danfei Xu},
  year = {2024},
  eprint = {2410.24221},
  archivePrefix = {arXiv},
  primaryClass = {cs.RO},
  url = {https://arxiv.org/abs/2410.24221}
}

@misc{zheng2026egoscalescalingdexterousmanipulation,
  title = {{EgoScale}: Scaling Dexterous Manipulation with Diverse Egocentric Human Data},
  author = {Ruijie Zheng and Dantong Niu and Yuqi Xie and Jing Wang and Mengda Xu and Yunfan Jiang and Fernando Casta{\~n}eda and Fengyuan Hu and You Liang Tan and Letian Fu and Trevor Darrell and Furong Huang and Yuke Zhu and Danfei Xu and Linxi Fan},
  year = {2026},
  eprint = {2602.16710},
  archivePrefix = {arXiv},
  primaryClass = {cs.RO},
  url = {https://arxiv.org/abs/2602.16710}
}

@misc{sharma2026demonstrationobjectsgeneralizingmanipulation,
      title={One Demonstration, Many Objects: Generalizing Manipulation via Local Contact Geometry}, 
      author={Satvik Sharma and Samrat Sahoo and Huang Huang and Fei-Fei Li and Jiajun Wu and Dorsa Sadigh and Jeannette Bohg},
      year={2026},
      eprint={2609.01938},
      archivePrefix={arXiv},
      primaryClass={cs.RO},
      url={https://arxiv.org/abs/2609.01938}, 
}

@misc{nair2022r3m,
      title={R3M: A Universal Visual Representation for Robot Manipulation}, 
      author={Suraj Nair and Aravind Rajeswaran and Vikash Kumar and Chelsea Finn and Abhinav Gupta},
      year={2022},
      eprint={2203.12601},
      archivePrefix={arXiv},
      primaryClass={cs.RO},
      url={https://arxiv.org/abs/2203.12601}, 
}

@misc{chen2026gapgraphaspolicymultiagentselflearning,
      title={GaP: A Graph-as-Policy Multi-Agent Self-Learning Harness For Variational Automation Tasks}, 
      author={Kaiyuan Chen and Shuangyu Xie and Letian Fu and Justin Yu and William Pacini and Sandeep Bajamahal and Hudson Kim and Jaimyn Drake and Daehwa Kim and Haoru Xue and Jonathan Francis and Christian Juette and Peter Schaldenbrand and Muhammet Yunus Seker and Ruwan Wickramarachchi and Uksang Yoo and Guanzhi Wang and Adithyavairavan Murali and Balakumar Sundaralingam and S. Shankar Sastry and Spencer Huang and Yuke Zhu and Linxi "Jim" Fan and Ken Goldberg},
      year={2026},
      eprint={2607.05369},
      archivePrefix={arXiv},
      primaryClass={cs.RO},
      url={https://arxiv.org/abs/2607.05369}, 
}

@misc{fu2026capxframeworkbenchmarkingimproving,
      title={CaP-X: A Framework for Benchmarking and Improving Coding Agents for Robot Manipulation}, 
      author={Letian Fu and Justin Yu and Karim El-Refai and Ethan Kou and Haoru Xue and Huang Huang and Wenli Xiao and Guanzhi Wang and Dantong Niu and Fei-Fei Li and Guanya Shi and Jiajun Wu and Shankar Sastry and Yuke Zhu and Ken Goldberg and Linxi "Jim" Fan},
      year={2026},
      eprint={2603.22435},
      archivePrefix={arXiv},
      primaryClass={cs.RO},
      url={https://arxiv.org/abs/2603.22435}, 
}

\clearpage

\section*{Appendix}

\subsection{Additional Robot Data Generation Examples}
We provide additional examples of how the verified robot trajectories are converted into data for policy training. For each task, the demonstrator is removed from the original egocentric video and replaced with the retextured robot and objects rendered in Blender. Left- and right-wrist camera views are generated in addition to the egocentric view, providing three synchronized views for each training sample.

To increase the diversity of the generated data, we first randomize the initial object states and sample 1,500 random seeds over the valid workspace on the table. For each sampled configuration, DexAgent checks whether the task remains feasible under the new object placement, including whether the required objects are reachable by the corresponding left or right hand and whether a valid execution can be found. Seeds that fail these feasibility checks are discarded. From the successful seeds, we randomly select 500 passed seeds with trajectories for policy training, with each configuration rendered from the egocentric, left-wrist, and right-wrist views. If fewer than 500 valid seeds are obtained from the initial 1,500 seeds, additional seeds are sampled and evaluated until 500 successful samples are collected. In this way, a single demonstrated task can be expanded into a larger set of physically feasible training episodes with varied object placements and viewpoints (\Cref{fig:appendix_inpainting}).

\subsection{Example Outputs from the Semantic Understanding Stage}
We provide example \texttt{scene\_semantics.json} files generated by Stage 1. These examples illustrate how DexAgent extracts task-relevant objects and their properties, decomposes the demonstration into an ordered sequence of subtasks, identifies the acting hand and its role in each subtask, and defines observable success conditions used to guide and verify subsequent stages.

\begin{strip}
\centering
\includegraphics[width=1\linewidth]{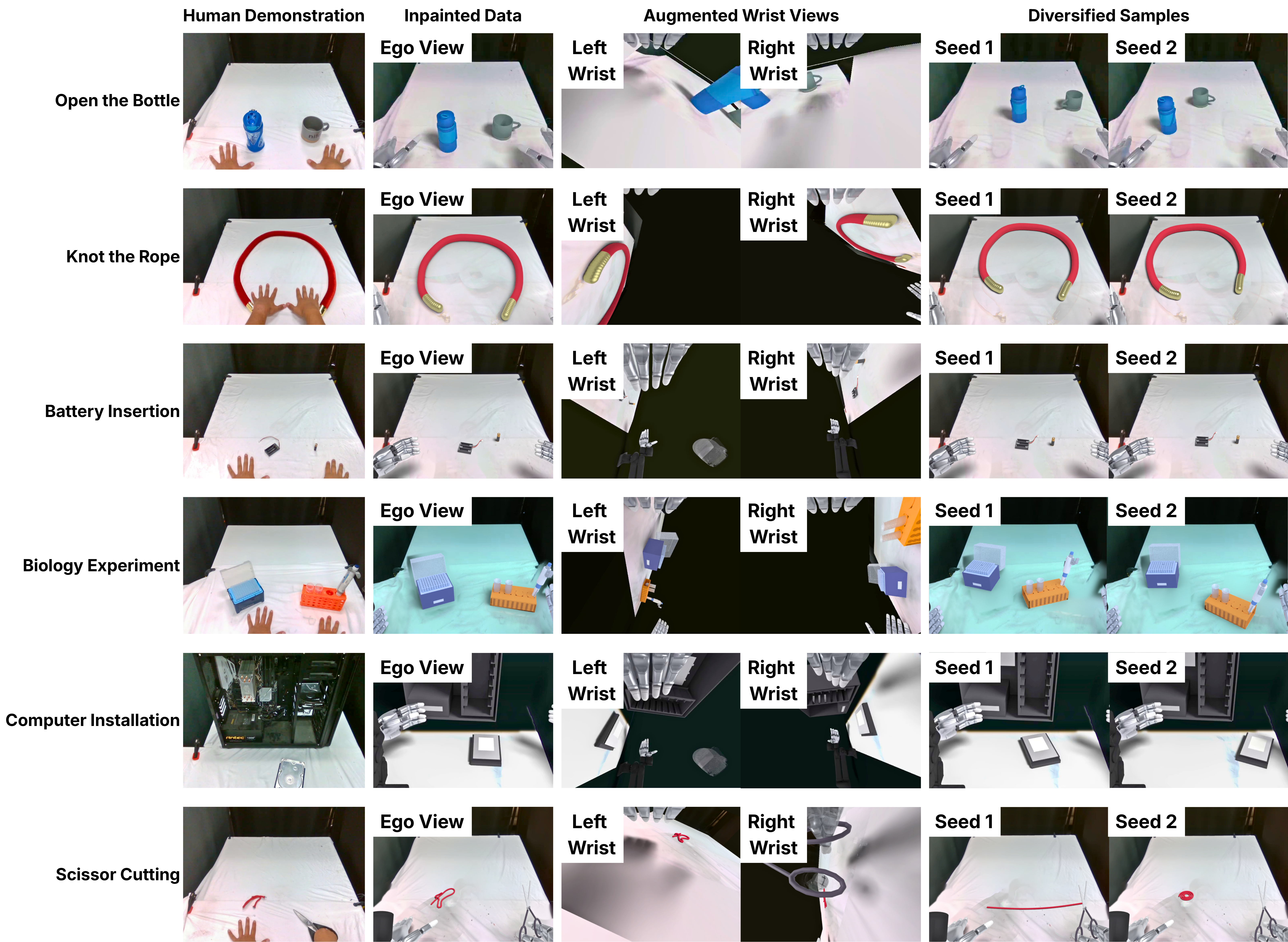}
\captionof{figure}{Examples of robot data generation in Stage 4. \textbf{Column 1} shows the original human demonstration. \textbf{Column 2} shows the egocentric view after removing the demonstrator and rendering the robot and objects into the scene, after being retextured in Blender. \textbf{Column 3-4} shows the corresponding generated left- and right-wrist camera views used for policy training. \textbf{Column 5-6} shows examples generated from the same demonstration with different random seeds, illustrating the data diversification happened during data generation.}
\label{fig:appendix_inpainting}
\end{strip}

\clearpage
\promptsec{Drawing and Wiping}
\begin{strip}
\begin{lstlisting}[basicstyle=\ttfamily\bsizeB, frame=single,
breaklines=true, columns=flexible]
## scene_semantics.json for drawing and wiping sample.

{
  ...,
  "task_summary": "Write 1, 2, 3, and 4 with the right hand; erase only 4 with the left hand; return both tools.",
  "objects": [
    {
      "id": "board", "name": "Framed whiteboard",
      "type": "rigid", ...,
      "properties": [
        ...,
        "Marker contact produces visible dark strokes that persist after the tip lifts.",
        "Sliding eraser contact removes contacted strokes while leaving unwiped strokes visible."
      ], ...
    },
    {
      "id": "marker", ...,
      "type": "rigid", ...,
      "properties": [
        ...,
        "Moving the tip against the board leaves a visible dark trace."
      ], ...
    },
    {
      "id": "eraser", ...,
      "type": "rigid", ...,
      "properties": [
        ...,
        "Its broad wiping face removes marker strokes through sliding contact with the board."
      ], ...
    }
  ],
  ...,
  "subgoals": [
    ...,
    {
      ...,
      "acting_hand": "right",
      "action": "Write digit 1.",
      ...,
      "hand_roles": {
        "left": "Holds the eraser at the left side.",
        "right": "Moves the marker tip along the stroke."
      },
      "goal": "A dark 1 is visible on the board."
    },
    ...,
    {
      ...,
      "acting_hand": "right",
      "action": "Write digit 4 to the right of 3.",
      ...,
      "goal": "The board displays 1, 2, 3, and 4 in order."
    },
    ...,
    {
      ...,
      "acting_hand": "left",
      "action": "Wipe repeatedly over digit 4.",
      ...,
      "hand_roles": {
        "left": "Slides the eraser back and forth over 4.",
        "right": "Holds the marker away from the board."
      },
      "goal": "Digit 4 is no longer visibly readable; 1, 2, and 3 remain."
    },
    ...
  ],
  ...
}
\end{lstlisting}
\end{strip}

\clearpage
\promptsec{Open the Bottle}
\begin{strip}
\begin{lstlisting}[basicstyle=\ttfamily\bsizeB, frame=single,
breaklines=true, columns=flexible]
## scene_semantics.json for open the bottle sample.

{
  ...,
  "task_summary": "Remove the screw cap, direct a pouring motion into the mug, and return the uncapped bottle upright.",
  "objects": [
    {
      "id": "bottle", ...,
      "type": "articulated",
      "parts": ["body", "mouth", "screw_cap", "attached_tube"],
      "properties": [
        ...,
        "The cap turns relative to the body before detaching from the threaded mouth.",
        ...,
        "A long tube is carried with the cap; it must clear the mouth before the cap is moved aside.",
        ...
      ], ...
    },
    {
      "id": "mug", ...,
      "type": "rigid", ...,
      "properties": [
        "Open-topped receptacle with a rigid body and a stable resting base.",
        ...,
        "Remains on the tabletop; neither hand grasps or stabilizes it."
      ], ...
    }
  ],
  ...,
  "subgoals": [
    ...,
    {
      ...,
      "acting_hand": "both",
      "action": "Loosen the cap with repeated turns.",
      ...,
      "hand_roles": {
        "left": "Resists rotation of the body.",
        "right": "Turns the cap and readjusts its grip as needed."
      },
      "goal": "Cap is loosened sufficiently for removal."
    },
    {
      ...,
      "acting_hand": "both",
      "action": "Lift off the cap assembly.",
      ...,
      "hand_roles": {
        "left": "Keeps the body upright and steady.",
        "right": "Lifts the cap and its attached tube clear."
      },
      "goal": "Cap is detached and the tube clears the bottle mouth."
    },
    ...,
    {
      ...,
      "acting_hand": "left",
      "action": "Tilt and hold the bottle in a pouring posture.",
      ...,
      "hand_roles": {
        "left": "Tilts the bottle while aiming its mouth into the mug.",
        "right": "Remains by the cap without manipulating the mug."
      },
      "goal": "Intended task outcome: water transfers from the bottle into the mug.",
      "observed_outcome": "Pouring posture is visible; a liquid stream or level change is not clearly resolved."
    },
    ...
  ],
  ...
}
\end{lstlisting}
\end{strip}

\clearpage
\promptsec{Place Toy in Drawer}
\begin{strip}
\begin{lstlisting}[basicstyle=\ttfamily\bsizeB, frame=single,
breaklines=true, columns=flexible]
## scene_semantics.json for place toy in drawer sample.

{
  ...,
  "objects": [
    {
      "id": "drawer_unit",
      "type": "articulated",
      "parts": ["housing", "sliding_tray", "front_handle"],
      "appearance": {
        "housing": "Gray",
        "front_panel": "Gray",
        "tray_interior": "Dark gray",
        "handle": "Black, front-mounted pull handle"
      },
      "properties": [
        "A single-DOF prismatic joint connects the tray to the housing.",
        "Bidirectional translation along the housing's longitudinal axis: outward to open and inward to close.",
        "Relative rotation and translation perpendicular to the sliding axis are constrained.",
        "Observed opening travel is approximately one housing depth; a rough visual estimate, not a measured mechanical travel limit."
      ],
      ...
    },
    {"id": "toy", "name": "Small yellow plush toy", ...},
    ...
  ],
  "subgoals": [
    ...,
    {
      "acting_hand": "left",
      "action": "Pull the drawer outward.",
      "object_refs": ["drawer_unit"],
      "goal": "Drawer interior and the free front placement region are accessible."
    },
    {
      "acting_hand": "right",
      "action": "Pick up the toy.",
      "object_refs": ["toy"],
      "goal": "Toy is lifted clear of the tabletop."
    },
    {
      "acting_hand": "right",
      "action": "Move the toy over the open drawer.",
      "object_refs": ["toy", "drawer_unit"],
      "goal": "Toy is above the drawer interior, clear of the existing contents."
    },
    {
      "acting_hand": "right",
      "action": "Release the toy into the drawer.",
      "object_refs": ["toy", "drawer_unit"],
      "goal": "Toy rests in the front region of the tray without hand support."
    },
    ...,
    {
      "acting_hand": "left",
      "action": "Push the drawer closed using its front handle.",
      "object_refs": ["drawer_unit"],
      "goal": "Front panel returns to the closed position with the toy enclosed."
    },
    ...
  ],
  ...
}
\end{lstlisting}
\end{strip}

\subsection{Additional Simulation Reconstruction and Robot Trajectory Optimization Examples}
\begin{figure*}[t]
\centering
\includegraphics[width=1\linewidth]{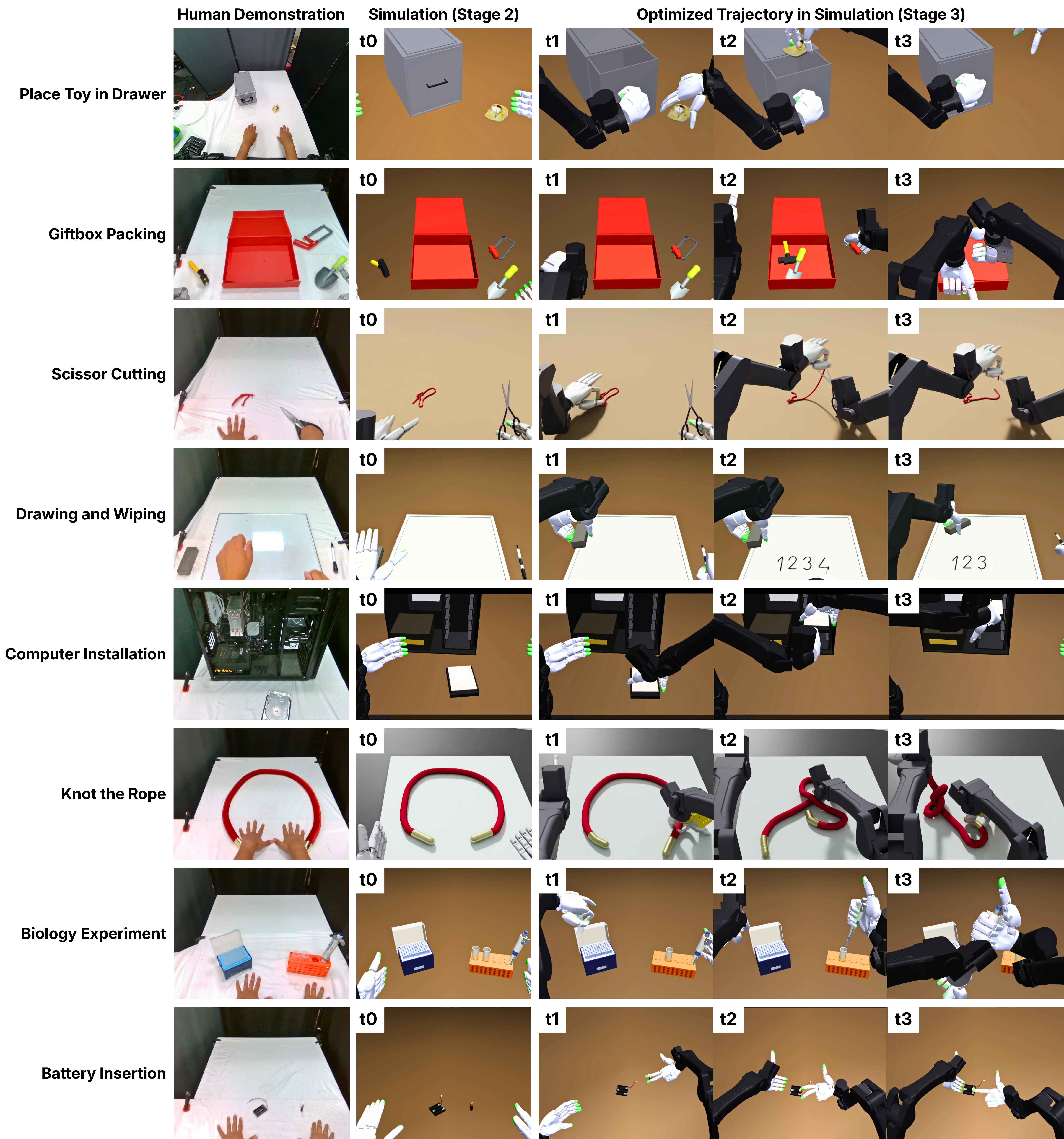}
\caption{Additional examples of property-based simulation reconstruction in Stage 2 and robot trajectory optimization in simulation in Stage 3. \textbf{Column 1} shows the recorded human demonstration, and \textbf{Column 2} shows the corresponding simulation scene reconstructed in Stage 2. \textbf{Column 3-5} show representative frames \(t_0\)–\(t_3\) from the robot trajectory optimized in simulation from Stage 3.}
\label{fig:appendix_simulation}
\end{figure*}

We provide additional examples of the simulation reconstruction and trajectory optimization stages on tasks not included in the main paper. These examples span rigid, articulated, and deformable objects, as well as a range of manipulation behaviors. Each row traces a single task from the recorded human demonstration, through the reconstructed simulation scene in Stage 2, to the optimized robot trajectory produced in Stage 3 (\Cref{fig:appendix_simulation}).

\clearpage
\begin{strip}
\centering
\includegraphics[width=1\linewidth]{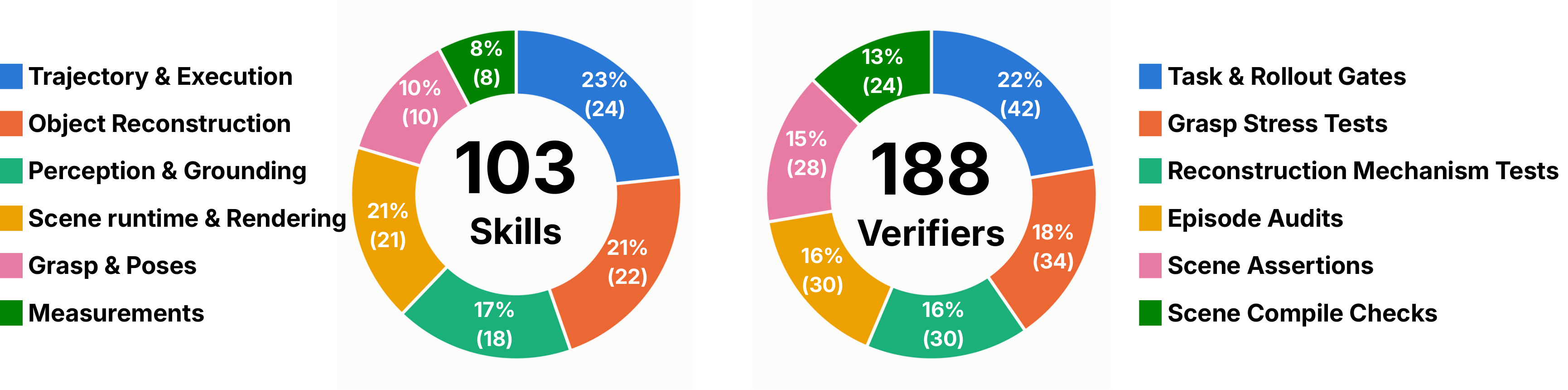}
\captionof{figure}{Breakdown of the current DexAgent tool library.
The library contains \textbf{103 skills} and \textbf{188 verifiers} grouped by functionality. \textit{Skills} cover perception, reconstruction, grasp and pose generation, scene execution, measurement, and trajectory generation, while \textit{verifiers} check task completion, grasp stability, reconstructed mechanisms, full-episode consistency, task-specific scene conditions, and scene validity. These skills and verifiers are expected to generalize to new and unseen samples, and the library is expected to continue to grow as DexAgent processes more new samples.}
\label{fig:appendix_library}
\end{strip}

\subsection{Current Library Tools Breakdown}

The current library contains 103 skills and 188 verifiers accumulated from the EgoDex experiment and our task samples. These skills and verifiers span from Stage 1 to Stage 3 in DexAgent, including scene semantics understanding tools, reconstruction tools, and trajectory optimization tools. Fig.~8 groups them into 6 categories each by function.

\textbf{Skills} include: \emph{Trajectory \& Execution}, which generates and connects robot motions (e.g., subgoal chaining and trajectory export); \emph{Object Reconstruction}, which builds task-relevant geometry and mechanisms (e.g., screw caps, drawers, and writable surfaces); \emph{Perception \& Grounding}, which extracts objects, depth, and human hand motion from video; \emph{Scene Runtime \& Rendering}, which compiles, executes, and renders simulation scenes; \emph{Grasp \& Poses}, which generates physically valid hand configurations; and \emph{Measurements}, which extracts quantities such as object motion and obstacle clearance.

\textbf{Verifiers} include: \emph{Task \& Rollout Gates}, which check task completion and final states; \emph{Grasp Stress Tests}, which evaluate grasp stability under lifting, shaking, and sustained loading; \emph{Reconstruction Mechanism Tests}, which validate simulated mechanisms such as threads, latches, and writable surfaces; \emph{Episode Audits}, which check consistency over the complete trajectory; \emph{Scene Assertions}, which verify task-specific conditions such as contact, IoU, and joint range; and \emph{Scene Compile Checks}, which detect structural failures such as penetration, unsupported objects, or missing mechanisms. 

These tools and verifiers are expected to be generalizable to new sample and new task, and newly developed tools are retained as DexAgent processes additional demonstrations, allowing the library to grow with reusable capabilities and checks.

\end{document}